\documentclass[lettersize,journal, twoside]{IEEEtran}
\usepackage{amsmath,amsfonts}
\usepackage{algorithmic}
\usepackage{algorithm}
\usepackage{array}
\usepackage[caption=false,font=normalsize,labelfont=sf,textfont=sf]{subfig}
\usepackage{textcomp}
\usepackage{stfloats}
\usepackage{url}
\usepackage{verbatim}
\usepackage{graphicx}
\usepackage{multirow} 
\usepackage{multicol}
\usepackage[hidelinks]{hyperref}
\usepackage{cite}
\usepackage{tikz,xcolor}
\usepackage{bm}
\usepackage{fancyhdr}
\usepackage{booktabs}
\usepackage{lettrine}
\usepackage{verbatim}
\usepackage{threeparttable}
\usepackage{makecell}
\usepackage{xcolor}
\usepackage[T1]{fontenc}
\usepackage{cite}
\usepackage{balance}

\usepackage{mathbbol}
\usepackage{amssymb} 
\usepackage[T1]{fontenc}
\DeclareUnicodeCharacter{0301}{\'{e}}
\usepackage{braket}
\usepackage{stmaryrd}

\def\x{{\mathbf x}}

\def\etal{{\em et al.}}

\definecolor{lime}{HTML}{A6CE39}
\DeclareRobustCommand{\orcidicon}{
	\begin{tikzpicture}
		\draw[lime, fill=lime] (0,0)
		circle[radius=0.16]
		node[white]{{\fontfamily{qag}\selectfont \tiny \.{I}D}}; \draw[white, fill=white] (-0.0625,0.095) circle [radius=0.007];
	\end{tikzpicture}
	\hspace{-2mm}
}
\foreach \x in {A, ..., Z}{%
	\expandafter\xdef\csname orcid\x\endcsname{\noexpand\href{https://orcid.org/\csname orcidauthor\x\endcsname}{\noexpand\orcidicon}}
}

\begin{document}

\title{Cross-Modal Information-Guided Network using Contrastive Learning for Point Cloud Registration}

\author{Yifan Xie\orcidA{}, Jihua Zhu\orcidC{}, Shiqi Li\orcidD{} and Pengcheng Shi\orcidB{}
\thanks{This work was supported in part by the National Key R\&D Program of China under Grant 2020AAA0109602, in part by the Key Research and Development Program of Shaanxi Province under Grant 2021GY-025 and Grant 2021GXLH-Z-097. (Corresponding author: Jihua Zhu.)}
\thanks{The authors are with the School of Software, Xi'an Jiaotong University, Xi'an 710000, China(e-mail:xieyifan@stu.xjtu.edu.cn; zhujh@xjtu.edu.cn; lishiqi@stu.xjtu.edu.cn; spcbruea@stu.xjtu.edu.cn). Code will be available at https://github.com/IvanXie416/CMIGNet.}
\thanks{Digital Object Identifier (DOI): see top of this page.}
}

\markboth{IEEE ROBOTICS AND AUTOMATION LETTERS. PREPRINT VERSION. NOVEMBER 2023}%
{ Xie \MakeLowercase{\textit{et al.}}: Cross-Modal Information-Guided Network using Contrastive Learning for Point Cloud Registration}


\maketitle

\begin{abstract}
The majority of point cloud registration methods currently rely on extracting features from points. However, these methods are limited by their dependence on information obtained from a single modality of points, which can result in deficiencies such as inadequate perception of global features and a lack of texture information. Actually, humans can employ visual information learned from 2D images to comprehend the 3D world. Based on this fact, we present a novel Cross-Modal Information-Guided Network (CMIGNet), which obtains global shape perception through cross-modal information to achieve precise and robust point cloud registration.
Specifically, we first incorporate the projected images from the point clouds and fuse the cross-modal features using the attention mechanism. Furthermore, we employ two contrastive learning strategies, namely overlapping contrastive learning and cross-modal contrastive learning. The former focuses on features in overlapping regions, while the latter emphasizes the correspondences between 2D and 3D features. Finally, we propose a mask prediction module to identify keypoints in the point clouds.
Extensive experiments on several benchmark datasets demonstrate that our network achieves superior registration performance.
\end{abstract}


\section{Introduction}
\IEEEPARstart{W}{ith} the rapid development of modern information technology and graphics, 3D reconstruction technology~\cite{ma2018review} has gained widespread application across various fields such as augmented reality~\cite{billinghurst2015survey}, simultaneous localization and mapping (SLAM)~\cite{ding2019deepmapping} and autonomous driving~\cite{geiger2012we}. This technology relies on point cloud information collected by scanning equipment from the surface of a target scene, which is then processed and reconstructed to form a 3D digital model of the scene.

One of the most important and challenging problems in the 3D reconstruction process is 3D point cloud registration~\cite{huang2021comprehensive}, which involves predicting a rigid 3D transformation and aligning the source point cloud with the target point cloud. The feature-awareness capability is crucial for accurately aligning the two sets of point clouds, especially in cases where the point clouds are partially occluded or contaminated with noise. As a result, improving the feature perception ability of point clouds has become a hot topic in the field of point cloud registration.

In the real world, humans possess an extraordinary capability to learn visual information from 2D images and apply this knowledge to understand the 3D world. For example, people can easily recognize a 3D object from a given 2D image. In practical applications such as robotics and autonomous driving, comprehending the 2D-3D correspondences would significantly enhance our ability to understand the 3D world. However, point clouds are a 3D spatial representation composed of sparse and disordered points, which distinguishes them from 2D images with dense and regular pixel arrays. Previous studies have tended to treat the understanding of 2D images and 3D point clouds as distinct problems. 
On the one hand, 2D images offer rich color and texture,  but they can be ambiguous in terms of depth and shape perception. On the other hand, point clouds provide crucial information on spatial and geometric details, but only capture local and texture-free features.

\begin{figure}
	\centering
	\includegraphics[width=0.95\columnwidth]{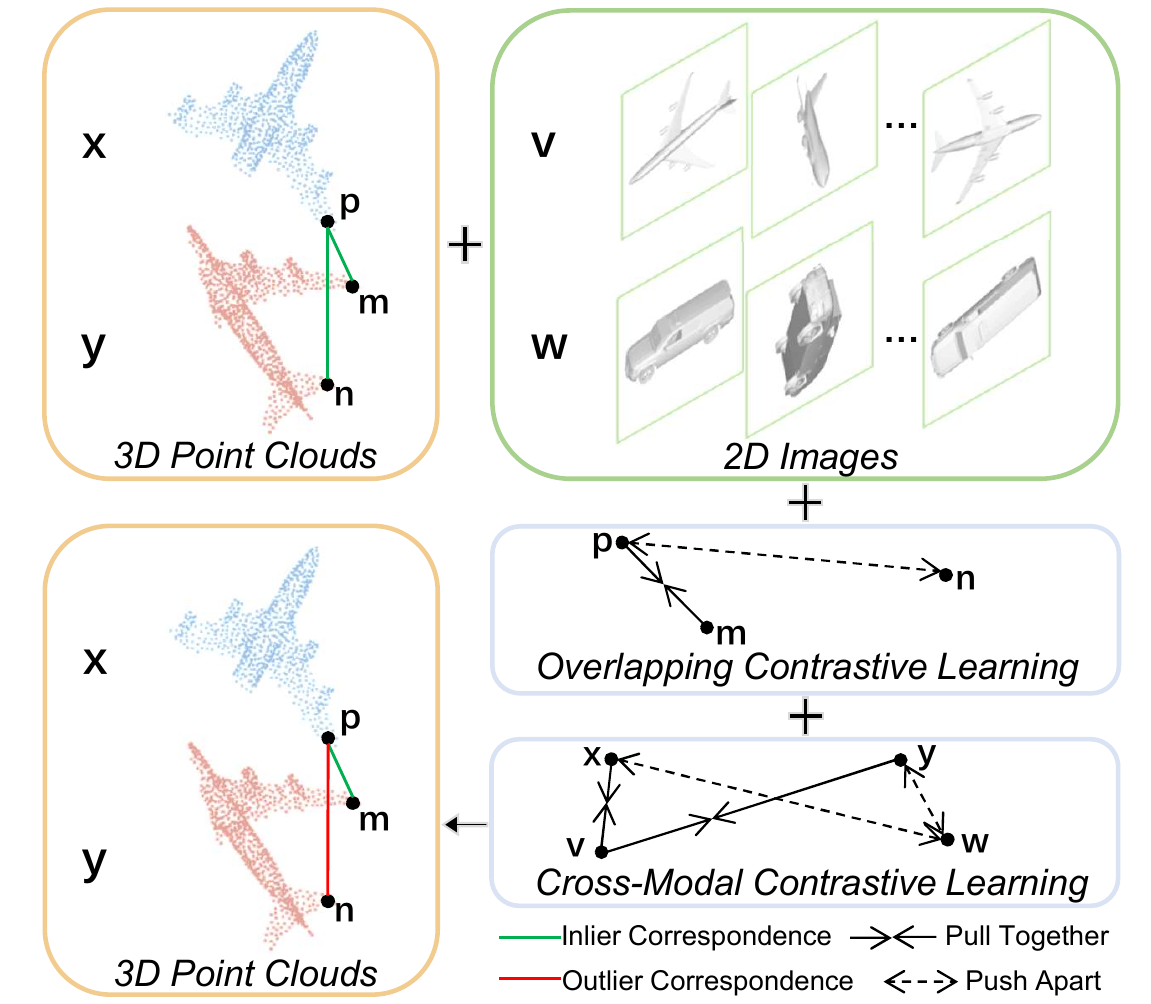} 
	\setlength{\belowcaptionskip}{-0.5cm}
	\caption{Most previous methods misclassify the correspondence of p and n as an inlier correspondence due to their similar local structures(top left). However, our method overcomes this issue by incorporating cross-modal image information and utilizing two contrastive learning strategies. As a result, the correspondence of p and n is correctly identified as an outlier correspondence(down left).
	}
	\label{figure1}
	\vspace*{-0.6\baselineskip}
\end{figure}

In this paper, our goal is to improve the point cloud registration problem by fusing image modality. 
We propose the Cross-Modal Information-Guided Network (CMIGNet), which integrates concepts from multimodal learning~\cite{ramachandram2017deep}, contrastive learning~\cite{chen2020simple} and attention mechanisms~\cite{vaswani2017attention} to correct inaccurate registration results, as depicted in Fig.~\ref{figure1}.
Specifically, We project the point clouds as images from various viewpoints and extract features from the point clouds and images separately. We then highlight overlapping point features using overlapping contrastive learning, establish 2D-3D correspondences through cross-modal contrastive learning, and employ attention mechanisms for information interaction. We also predict the keypoints to minimize the negative impact of non-critical points on registration tasks. Finally, we use spatial coordinates and hybrid features to guide the search for correspondences independently, and extract rigid transformation according to singular value decomposition (SVD).

To summarize, the contributions of our paper include: 
\begin{itemize}
	
\item We propose a novel cross-modal point cloud registration network CMIGNet, which perceives the global shape to achieve more accurate registration. 
\item Our proposed method utilizes two contrastive learning strategies. The first is overlapping contrastive learning, which emphasizes the features of overlapping points. The second is cross-modal contrastive learning, which establishes 2D-3D correspondences.  
\item A new method for predicting point cloud masks is proposed to extract keypoints and reduce the consumption of computational resources.
\item Extensive experimental results on the various benchmark datasets demonstrate that our method can achieve superior registration performance.

\end{itemize}

\section{RELATED WORK}
\subsection{Point Cloud Registration}

Point cloud registration is a process that aims to transform and align input point clouds with each other. 
The Iterative Closest Point (ICP) algorithm~\cite{besl1992method} is a widely used rigid alignment method that iteratively optimizes the distance between two point clouds to maximize their overlap. However, the ICP algorithm has some limitations, including its reliance on initial poses and sensitivity to outlier points. Consequently, many variants of the ICP algorithm have been developed to address these issues.
One such variant is Go-ICP~\cite{yang2015go}, which employs a branch-and-bound approach to search for the globally optimal registration result at the expense of longer computation time. 

With the exceptional results demonstrated by deep learning in image processing, researchers have turned their attention to learning-based point cloud registration methods.
PointNetLK~\cite{aoki2019pointnetlk} merges a modified Lucas Kanade algorithm~\cite{baker2004lucas} into the PointNet~\cite{qi2017pointnet} to iteratively align the input point clouds.
DCP~\cite{wang2019deep} combines DGCNN~\cite{wang2019dynamic} and attention modules~\cite{vaswani2017attention} to extract features and uses pointer networks to predict soft matches between point clouds. To further tackle the partial overlap problem, PRNet~\cite{Wang_Solomon_2019} utilizes keypoint detection to select the common points of the input point clouds.
MaskNet~\cite{sarode2020masknet} introduces a fully convolutional neural network that identifies the most similar points in one point cloud to those in another.
IDAM~\cite{li2020iterative} develops a two-stage point elimination technique to help generate partial correspondences.
OMNet~\cite{xu2021omnet} is capable of learning overlap masks, which it uses to identify non-overlapping regions. FINet~\cite{xu2022finet} utilizes a two-branch structure that allows it to handle rotations and translations separately, and it also enhances the correlation information between inputs at multiple stages of the registration process.
VRNet~\cite{zhang2022vrnet} introduces a new class of virtual points named rectified virtual corresponding points. These points have the same shape as the source point cloud and the same pose as the target point cloud.
While previous approaches have focused on the matching phase, our work introduces cross-modal information and prioritizes feature interactions.

\subsection{Cross-Modal Learning}

Cross-modal learning aims to increase the diversity of data by leveraging information from multiple modalities to improve the performance and generalization of the model.
A number of representative cross-modal learning approaches have emerged.
For example, CLIP~\cite{radford2021learning} learns multimodal embedding spaces by maximizing the cosine similarity between image and text modalities.
Afham \etal~\cite{afham2022crosspoint} employed unsupervised methods to encourage the embedding of 2D image features closer to 3D point cloud prototypes. 
PointCMT~\cite{yan2022let} is the pioneering approach to conduct knowledge distillation from image-to-point for point cloud analysis.
IMFNet~\cite{huang2022imfnet} uses cross-modal features for point cloud registration on real datasets.
Compared to existing methods, our method applies a cross-modal feature correspondence method based on contrastive learning and uses an attention mechanism to fuse 2D and 3D features, forming an end-to-end point cloud registration network framework.

\begin{figure*}
	\centering
	\includegraphics[width=0.95\textwidth]{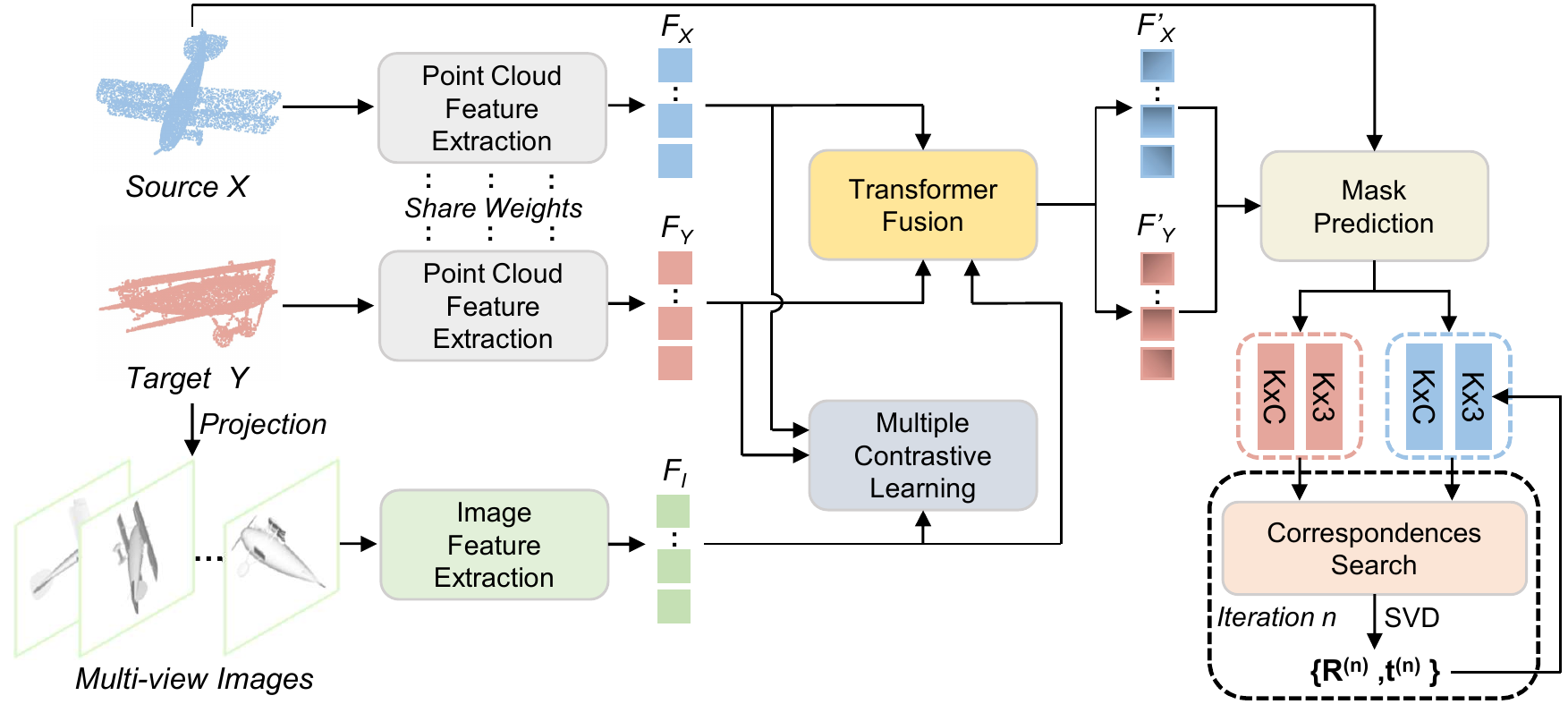} 
	\caption{The overall architecture of our CMIGNet. Given point clouds ${X}$, ${Y}$ and projective multi-view images, we first extract the point cloud features and image features separately. Then, Multiple Contrastive Learning highlights the features of overlapping points and establishes 2D-3D correspondences, while Transformer Fusion facilitates information interaction. Further, Mask Prediction identifies keypoints in the point cloud. Finally, Correspondences Search determines the final matching matrix, which is then used to estimate the rigid transformation $\{\textbf{R},\textbf{t}\}$ by the SVD method. CMIGNet achieve accurate registration by iteration of Correspondences Search and SVD estimation. Before next iteration, the current rigid transformation is utilized to transform the source point cloud $X$ into a new position for new Correspondences Search.}
	\label{figure2}
	\vspace{-.3cm}
\end{figure*}

\section{OUR METHOD}

Given two unaligned point clouds $X$ and $Y$,
where $X=\{x_1, \ldots, x_i, \ldots, x_N\}$ and $Y=\{y_1, \ldots, y_j, \ldots, y_M\}$.
our objective is to find the rigid transformation $\{\textbf{R}, \textbf{t}\}$ to align the two point clouds, where $\textbf{R} \in S O(3)$ is a rotation matrix and $\textbf{t} \in \mathbb{R}^3$ is a translation vector.
The one-to-one correspondence between points is not required in our method, which means $ N \neq M $ in most cases.
Fig.~\ref{figure2} shows the architecture of our CMIGNet.

\subsection{Feature Extraction}

The feature extraction module is divided into two parts, which are point cloud feature extraction and image feature extraction.
For point cloud feature extraction, we treat each point in the point clouds $X$ and $Y$ as a vertex in a graph. Then, we can calculate the pointwise feature using the EdgeConv~\cite{wang2019dynamic} operation.
To broaden the perceptual field of vertices, we utilize the k-nearest neighbor (kNN) algorithm for graph construction at each layer. Simultaneously, we employ channel connectivity to enhance the fusion of features across different layers. 

For image feature extraction, we project 3D point cloud objects onto various viewpoints, resulting in $V$ corresponding 2D images. Thus we can obtain the final image features $F_{I} \in \mathbb{R}^{N \times C}$:
\begin{equation}
F_{I}=\mathcal{R}(\mathcal{A}\left\{\operatorname{CNN}\left(\mathcal{I}_v\right)\right\}_{v=1}^V).
\end{equation}
where $\mathcal{R}(\cdot)$ is the repeat operation, $\mathcal{A}\{\cdot\}$ is the aggregation function and $\mathcal{I}$ denotes the projected image.


\subsection{Multiple Contrastive Learning}
\emph{Overlapping contrastive learning.} 
We propose overlapping contrastive learning to highlight the features of overlapping regions and reduce the influence of non-overlapping points. 
Specifically, we apply a ground truth transformation to the source point cloud $X$. The points in the transformed point cloud are considered overlapping points if their minimum distance from the target point cloud $Y$ is less than the threshold value.
Then, we generate overlapping point features
${\mathcal{F}_{X}}=\left\{{f}_{X_i} \in \mathbb{R}^C\right\}_{i=1}^{M}$ and ${\mathcal{F}_{Y}}=\left\{{f}_{Y_i} \in \mathbb{R}^C\right\}_{i=1}^{N}$, 
as well as non-overlapping point features
${\mathcal{F}_X^{\prime}}=\left\{{{f}_{X_i}}^{\prime} \in \mathbb{R}^C\right\}_{i=1}^{M^{\prime}}$ and ${\mathcal{F}_Y^{\prime}}=\left\{{{f}_{Y_i}}^{\prime} \in \mathbb{R}^C\right\}_{i=1}^{N^{\prime}}$ through the overlap selection module.
We consider pairs of overlapping point features between the two point clouds as the positive pair set $\mathcal{P}$, pairs of overlapping point features of point cloud $X$ and non-overlapping point features of point cloud $Y$ as the negative pair set $\mathcal{N}_{1}$, and pairs of overlapping point features of point cloud $Y$ and non-overlapping point features of point cloud $X$ as the negative pair set $\mathcal{N}_{2}$. 
Based on this, our overlapping contrastive learning loss $\mathcal{L}_{OCL}$ can be constructed as follows:
\begin{equation}
    \footnotesize
    \label{eq:ocl()}
    \begin{aligned}
        &\mathcal{L}_{OCL} = \sum_{(i, j) \in \mathcal{P}}\left[D\left({f}_{X_i}, {f}_{Y_j}\right)-\sigma_p\right]_{+}^2 /|\mathcal{P}|
        \\&+\sum_{(i, j) \in \mathcal{N}_{1}}\left[\sigma_n-D\left({{f}_{X_i}}, {{f}_{Y_j}}^{'}\right)\right]_{+}^2 /|\mathcal{N}_{1}| 
        \\&+\sum_{(i, j) \in \mathcal{N}_{2}}\left[\sigma_n-D\left({{f}_{Y_i}}, {{f}_{X_j}}^{'}\right)\right]_{+}^2 /|\mathcal{N}_{2}|,
    \end{aligned}
\end{equation}
where $D(\cdot,\cdot)$ denotes the Euclidean distance between features and $[\cdot]_{+}$ represents a clamp function $\emph{max} (x, 0)$. $\sigma_p$ and $\sigma_n$ are margins for positive and negative pairs, which prevent the network from overfitting.

\begin{figure*}
	\centering
	\includegraphics[width=0.95\textwidth]{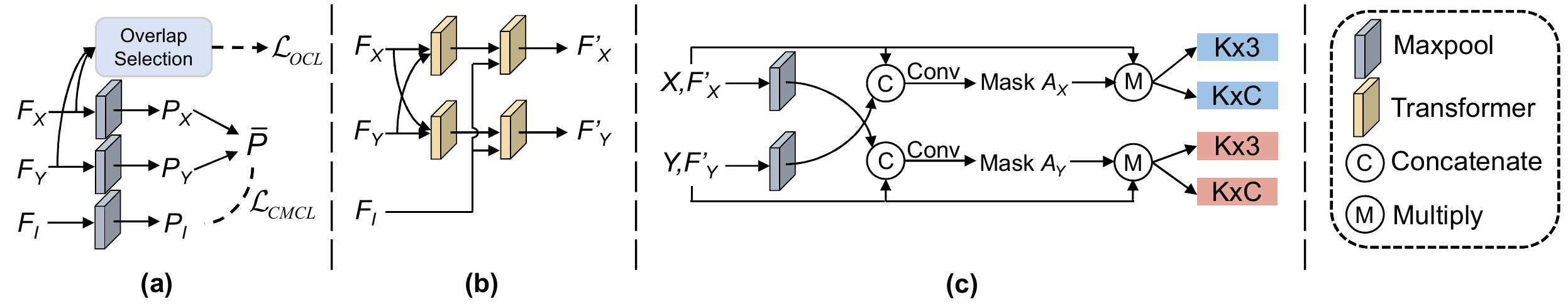} 
	\caption{The detailed structure of the main modules. (a) Multiple Contrastive Learning, (b) Transformer Fusion, and (c) Mask Prediction.}
	\label{figure3}
\end{figure*}

\emph{Cross-modal contrastive learning.} 
Cross-modal contrastive learning is utilized to establish 2D-3D correspondences.
As described in Fig.~\ref{figure3}(a), we utilize a pooling operation to project the point cloud features ${F}_{X}$ and ${F}_{Y}$, as well as the image features ${F}_{I}$, into the invariant space $\mathbb{R}^C$. This results in the projection vectors ${P}_{X}$, ${P}_{Y}$, and ${P}_{I}$. Then we calculate the average of ${P}_{X}$ and ${P}_{Y}$, which yields the projection vector ${\bar P}$ for the point cloud modality.

In the invariant space, our objective is to maximize the similarity between ${\bar P}$ and ${P}_{I}$, as they both correspond to the same object. 
Therefore, we construct positive samples $pos$:
\begin{equation}
\label{eq:pos()}
    pos = 
    {\exp(sim({\bar P}^b, {P}^b_{{I}})/\tau)},
\end{equation}
where $\tau$ is the temperature factor, $sim (\cdot,\cdot)$ denotes the cosine similarity function and $b$ is the serial number in the mini-batch.

We also aim to minimize the similarity between ${\bar P}$ and all other projected vectors in the mini-batch of point clouds and images. Thus, we construct negative samples $neg$:
\begin{equation}
\footnotesize
\label{eq:neg()}
    neg = 
   { \sum\limits_{\substack{k=1 \\ k \neq b}}^{N} \exp(sim({\bar P}^b, {\bar P}^k)/\tau) + \sum\limits_{\substack{k=1 \\ k \neq b}}^{N} \exp(sim({\bar P}^b, {P}^k_{{I}})/\tau)},
\end{equation}

where $N$ is the mini-batch size. $sim (\cdot,\cdot)$, $\tau$ and $b$ refer to the same parameters as in Eq. \ref{eq:pos()}.

Combining contrastive learning ideas, we compute the loss function $l\left(b, {\bar P}^b, {P}^b_{{I}}\right)$ as:
\begin{equation}
\label{eq:l()}
    l\left(b, {\bar P}^b, {P}^b_{{I}}\right) = -\log
    \frac{pos}{neg},
\end{equation}
and the cross-modal contrastive learning loss $\mathcal{L}_{CMCL}$ for a mini-batch is then formulated as:
\begin{equation}
\label{eq:cmcl()}
    \mathcal{L}_{CMCL}=\frac{1}{2 N} \sum_{i=1}^N[ l\left(b, {\bar P}^b, {P}^b_{{I}}\right)
    + l\left(b, {P}^b_{{I}}, {\bar P}^b\right)].
\end{equation}

\subsection{Transformer Fusion}

Given point cloud features and image features, two Transformer layers are employed to further extract contextual information.
The first Transformer layer is utilized to facilitate information interaction between point clouds, the input comprises ${F}_{X}$ and ${F}_{Y}$. After information interaction, we can obtain the point cloud interaction features $\Phi_{{X}}$ and $\Phi_{{Y}}$, which highlight the parts of keypoints.

\begin{figure}
	\centering
	\includegraphics[width=0.45\textwidth]{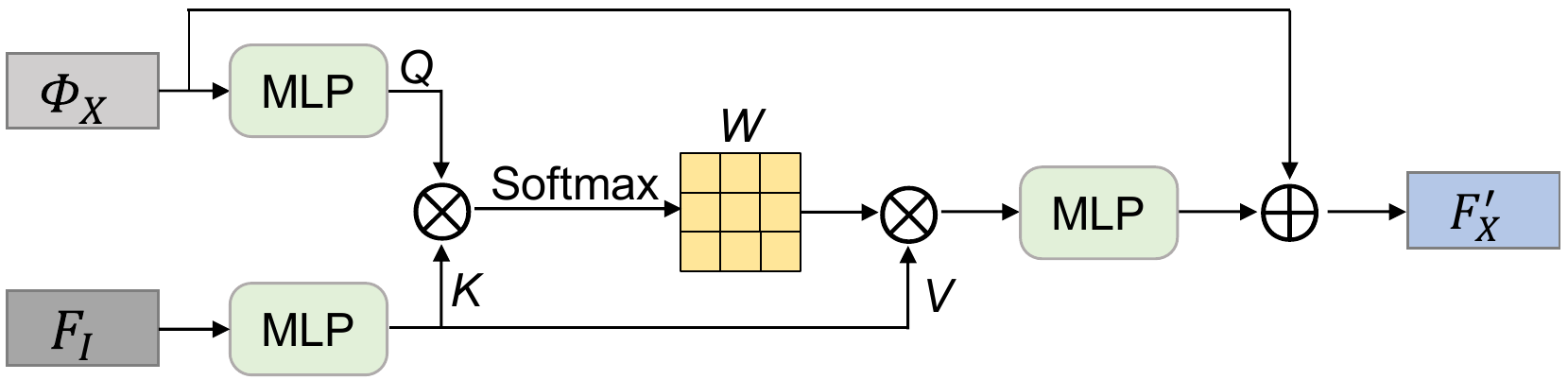} 
	\caption{The pipeline of the second Transformer layer in the Transformer Fusion module.
	}
	\label{figure4}
	\vspace{-.2cm}
\end{figure}

The purpose of the second Transformer layer is to enhance the distinctiveness of pointwise features by extracting global shape and texture information. As shown in Fig.~\ref{figure4}, taking the source point cloud $X$ as an example, the first step involves processing $\Phi_{{X}}$ and ${F}_{I}$ through an MLP. The output of $\Phi_{{X}}$ is treated as the query array $Q \in \mathbb{R}^{N \times C_t}$, while the output of ${F}_{I}$ is treated as the key array $K \in \mathbb{R}^{N \times C_t}$ and value array $V \in \mathbb{R}^{N \times C_t}$. The MLP output dimension is represented by $C_t$. The $W \in R^{N \times N}=\operatorname{softmax}\left(\frac{Q K^T}{\sqrt{C_t}}\right)$ represents the weight attributed to global shape and texture information that could aid in describing pointwise features. Then, we can calculate the final hybrid features ${F}_{X}^{\prime} \in \mathbb{R}^{N \times C}$:
\begin{equation}
\label{eq:l2()}
    {F}_{X}^{\prime} = \Phi_{{X}}+MLP(W\cdot V).
\end{equation}
Similarly, we can get ${F}_{Y}^{\prime}$ in the same way.

\subsection{Mask Prediction}
As illustrated in Fig.~\ref{figure3}(c), we propose a mask prediction module, which helps preserve the discriminative features while eliminating the non-discriminative ones.
Given the hybrid features ${F}_{X}^{\prime} \in \mathbb{R}^{N \times C}$ and coordinates ${X} \in \mathbb{R}^{N \times 3}$, we begin by pooling the features. Then we repeat the resulting pooled vector and concatenate it with the hybrid feature ${F}_{Y}^{\prime}$ of the target point cloud Y. This is followed by a one-dimensional convolution that yields a significance score for each feature. A higher significance score indicates that the feature is more discriminative, which is advantageous for the matching point search. 
Finally, we create the final mask ${A}_{X} \in \mathbb{R}^{N \times 1}$ by setting the mask of the $K$ points with the highest significance score to 1 and the mask of the remaining points to 0. This mask is then used to select the coordinates $({P}_{X} ^{k}$, ${P}_{Y} ^{k})$ and features $({F}_{X} ^{k}$, ${F}_{Y} ^{k})$ of the $K$ keypoints, which guide the subsequent search for correspondences.

\subsection{Correspondences Search}
\begin{figure}
	\centering
	\includegraphics[width=0.45\textwidth]{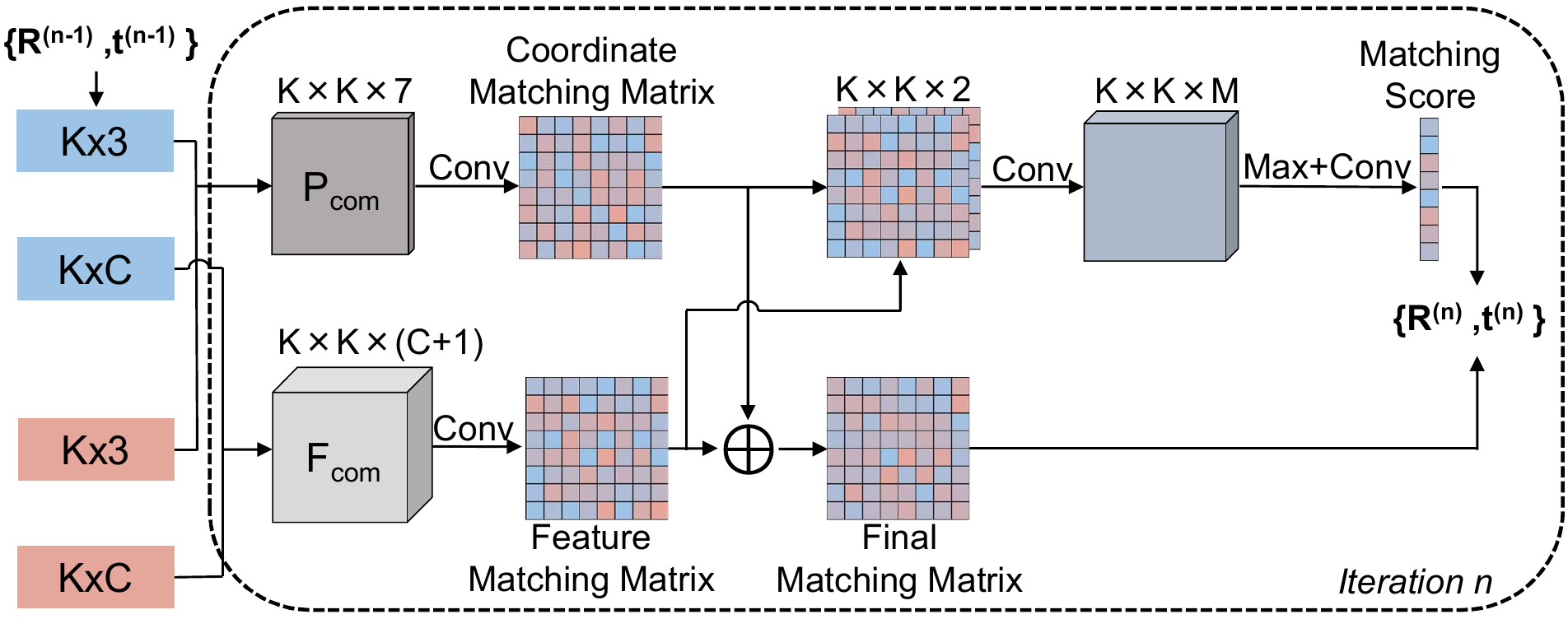} 
	\caption{The flow of the Correspondences Search module.
	}
	\label{figure5}
	\vspace{-.2cm}
\end{figure}

We propose that the hybrid features and spatial coordinates can be used to guide correspondences search independently. The entire process flow is illustrated in Fig.~\ref{figure5}.

Given keypoints' spatial coordinates ${P}_{X} ^{k}$, ${P}_{Y} ^{k}$ and hybrid features ${F}_{X} ^{k}$, ${F}_{Y} ^{k}$, we can form a combination of spatial coordinates and a combination of hybrid features. These combinations are compressed into one dimension to obtain coordinate matching matrix $M_P$ and feature matching matrix $M_F$. Then we add $M_P$ and $M_F$ to obtain the final matching matrix. We also obtain the matching score $s(i)$ of ${x}_i$ by performing maximum aggregation and convolution operations. Therefore the weight for the $i_{th}$ point pair is defined as:
\begin{equation}
w_i = \frac{s(i) \cdot \mathbb I \llbracket{s(i) \geq \text{median}_{k}(s(k))}\rrbracket}{\sum_i s(i) \cdot \mathbb I \llbracket{s(i) \geq \text{median}_{k}(s(k))}\rrbracket},
\end{equation}
where $\mathbb I \llbracket{\cdot}\rrbracket$ is the indicator function and $s(i)$ denotes the matching score of ${x}_i$. With this weight vector, we finally use the weighted SVD~\cite{arun1987least} to solve for the transformation matrix:
\begin{equation}
    \mathbf{R}, \mathbf{t}=\underset{\mathbf{R}, \mathbf{t}}{\operatorname{argmin}} \sum_i w_i\left\|\mathbf{R} {x}_i+\mathbf{t}-{x}_i^{\prime}\right\|^2,
\end{equation}
where ${x}_i^{\prime}$ is the corresponding point found by ${x}_i$ according to the final matching matrix.

\subsection{Loss Function}
\emph{Overlapping Contrastive Learning Loss.}
We utilize overlapping contrastive learning loss, denoted as $\mathcal{L}_{OCL}$, to highlight the features of overlapping regions. The loss can be found in Equation~\ref{eq:ocl()}.

\emph{Cross-Modal Contrastive Learning Loss.}
We utilize a cross-modal contrastive learning loss $\mathcal{L}_{CMCL}$ to minimize the distance between the 3D point cloud and its corresponding 2D image in the feature space. The specific formula for this loss function can be found in Equation~\ref{eq:cmcl()}.

\emph{Mask Prediction Loss.}
Despite the unavailability of direct keypoint annotations, we employ mutual-supervision loss~\cite{li2020iterative} to train our network. 
The underlying idea is that keypoints exhibit low entropy as they are confident in matching. As such, we define the loss for mask prediction as follows:
\begin{equation}
\footnotesize
    \mathcal{L}_{MP}=\frac{1}{K} \sum_{i=1}^K\left(a(i)-\sum_{j=1}^K M(i, j) \log (M(i, j))\right)^2,
\end{equation}
where $a(i)$ is the mask of ${x}_i$ and $M$ is the final matching matrix.

\emph{Matching Score Loss.}
The loss of matching score computation for the $n_{th}$ iteration is defined as:
\begin{equation}
\footnotesize
   \mathcal{L}_{MS}^{(n)}=\frac{1}{K} \sum_{i=1}^K-\hat{s}_i \log (s(i))-\left(1-\hat{s}_i\right) \log (1-s(i)),
\end{equation}
where $s(i)$ is the matching score of ${x}_i$ and $\hat{s}_i$ is the label indicating whether the correspondence distance under ground truth transformation is less than the distance threshold.

\emph{Correspondences Search Loss.}
The correspondences search loss is used to supervise the final matching matrix, and for the $n_{th}$ iteration, it is defined as:
\begin{equation}
\footnotesize
   \mathcal{L}_{CS}^{(n)}=\frac{1}{K} \sum_{i=1}^K-\hat{y}_i \log \left(M^{(n)}\left(i, j^*\right)\right),
\end{equation}
where $j^*$ refers to the index of the point closest to $x_i$ under ground truth transformation, and $\hat{y}_i$ is the label used to determine if the distance between ${x}_i$ and ${y}_{j^*}$ is less than the distance threshold.

The overall loss is the sum of the five losses:
\begin{equation}
\footnotesize
   \mathcal{L}_{total}=\mathcal{L}_{OCL} + \mathcal{L}_{CMCL} +\mathcal{L}_{MP}+\sum_n\left(\mathcal{L}_{MS}^{(n)}+\mathcal{L}_{CS}^{(n)}\right)
\end{equation}

\section{EXPERIMENTS}

\subsection{Experimental Settings}
\emph{Datasets.} We evaluate our method on ModelNet40~\cite{wu20153d}, Stanford 3D Scan~\cite{curless1996volumetric} and 7Scenes~\cite{shotton2013scene}. The ModelNet40 comprises 12,311 CAD models from 40 object categories. We use 9,843 models for training and 2,468 models for testing. The Stanford 3D Scan consists of 10 actual scans, and we reduced the size of each model in our trials by downsampling them to 10,000 points. The 7Scenes is a widely used benchmark for registration in indoor environments, comprising 7 scenes, namely \emph{Chess}, \emph{Fires}, \emph{Heads}, \emph{Office}, \emph{Pumpkin}, \emph{RedKitchen}, and \emph{Stairs}. The dataset is divided into 296 and 57 samples for training and testing.

\emph{Compared methods and evaluation metrics.}
We compare our method with traditional method ICP~\cite{besl1992method} and the learning-based methods, including PointNetLK~\cite{aoki2019pointnetlk}, DCP~\cite{wang2019deep}, PRNet~\cite{Wang_Solomon_2019}, IDAM~\cite{li2020iterative}, OMNet~\cite{xu2021omnet}, FINet~\cite{xu2022finet}, and VRNet~\cite{zhang2022vrnet}. We use the implementations of ICP in Intel Open3D~\cite{zhou2018open3d} and the others released by their authors.
Following~\cite{wang2019deep}, we measure anisotropic errors, including root mean squared error (RMSE) and mean absolute error (MAE) of rotation and translation.

\emph{Implementation Details.}
We train our network end-to-end using PyTorch implementation with 3090 GPU. We run 3 iterations during training and testing. We train our network with the Adam~\cite{kingma2014adam} optimizer for 100 epochs. The initial learning rate is $10^{-4}$ and is multiplied by 0.5 at 50 and 75 epochs.


\subsection{Evaluation on ModelNet40}

\emph{Same categories.}
We randomly selected 1,024 points from the outer surface of each model and applied rotations by sampling three Euler angle rotations within the $\left[0^{\circ},45^{\circ}\right]$ range, as well as translations within the $\left[-0.5,0.5\right]$ range, on each axis during both training and testing. We transform the source point cloud $X$ using the sampled rigid transform and the task is to register it to the unperturbed reference point cloud $Y$. To simulate partial-to-partial registration, we follow PRNet~\cite{Wang_Solomon_2019} to remove 25$\%$ points from both point clouds. From Table~\ref{table1}, one can see that our method obtains the lowest error among the traditional and learning-based methods. Example results are shown in Fig.~\ref{figure6}(a).

\begin{table}[]
    \small
    \centering
    \setlength\tabcolsep{8pt} 
    \caption{The registration result on same categories in ModelNet40.}
    \resizebox{\linewidth}{!}{%
        \begin{tabular}{lcccccc}
        
            \toprule
            Method     &RMSE(\textbf{R})       &MAE(\textbf{R})  &RMSE(\textbf{t})  &MAE(\textbf{t})        
            \\ 
            \midrule
            ICP~\cite{besl1992method}       & 33.684& 25.053& 0.2912& 0.2524 \\
            PointNetLK~\cite{aoki2019pointnetlk} &16.788 & 7.552&0.0429 &0.0289 \\
            DCP~\cite{wang2019deep} &6.649 & 4.847 & 0.0273 & 0.0215 \\
            PRNet~\cite{Wang_Solomon_2019} &  3.142&  1.458& 0.0163& 0.0119 \\
            IDAM~\cite{li2020iterative}&2.461&0.561 &0.0167 &\underline{0.0035} \\
            OMNet~\cite{xu2021omnet} &  1.499& 0.655& 0.0110&  0.0067 \\
            FINet~\cite{xu2022finet} & 1.463& 0.642& 0.0112 &0.0068 \\ 
            VRNet~\cite{zhang2022vrnet}& \underline{0.982}&  \underline{0.496}& \underline{0.0061}& 0.0039 \\
            \midrule
            Ours & \textbf{0.772}& \textbf{0.408}& \textbf{0.0048}&\textbf{0.0030}
            \\ 
            \bottomrule
        \end{tabular}%
    }
    \vspace{-0.3cm}
    \label{table1}
\end{table}

\emph{Unseen categories.}
In this experiment, we assess the generalization ability of our approach to unseen categories. Specifically, we evaluate its performance on 20 new categories that have not been previously seen by the model. To ensure a fair comparison, the data pre-processing steps used in this experiment are the same as those employed in the first experiment. Despite the new challenge presented by the unseen categories, our approach continues to yield excellent results. Table~\ref{table2} summarizes the results, and some of the visualization outcomes are presented in Fig.~\ref{figure6}(b).

\begin{table}[]
    \small
    \centering
    \setlength\tabcolsep{8pt} 
    \caption{The registration result on unseen categories in ModelNet40.}
    \resizebox{\linewidth}{!}{%
        \begin{tabular}{lcccccc}
        
            \toprule
            Method     &RMSE(\textbf{R})       &MAE(\textbf{R})  &RMSE(\textbf{t})  &MAE(\textbf{t})        
            \\ 
            \midrule
            ICP~\cite{besl1992method}       & 34.274& 25.637&  0.2924&0.2519 \\
            PointNetLK~\cite{aoki2019pointnetlk} & 22.824&  9.548&0.0621& 0.0214 \\
            DCP~\cite{wang2019deep} &9.837& 6.645&0.0338&  0.0252 \\
            PRNet~\cite{Wang_Solomon_2019} & 4.992&2.547&  0.0287&0.0149 \\
            IDAM~\cite{li2020iterative}&3.042&0.616& 0.0197&0.0048 \\
            OMNet~\cite{xu2021omnet} &2.625&1.010&0.0143& 0.0075 \\
            FINet~\cite{xu2022finet}  &2.391&0.801 & 0.0105&0.0045 \\ 
            VRNet~\cite{zhang2022vrnet}&\underline{2.121}& \underline{0.585}&\underline{0.0063}&\underline{0.0039} \\
            \midrule
            Ours & \textbf{0.842}& \textbf{0.431}& \textbf{0.0046}& \textbf{0.0029}
            \\ 
            \bottomrule
        \end{tabular}%
    }
    \vspace{-0.3cm}
    \label{table2}
\end{table}

\emph{Gaussian noise.}
Additionally, we assess our model's performance in the presence of noise, as it is commonly found in real-world point clouds. Similar to the first experiment, we apply the same preprocessing steps, but this time we introduce random Gaussian noise with a standard deviation of 0.01, clipped to $\left[-0.05,0.05\right]$, to all the point clouds. Table~\ref{table3} demonstrates that our method outperforms all other approaches. Furthermore, Fig.~\ref{figure6}(c) displays some example results.

\emph{Gaussian Noise with Lower Overlap.}
Finally, in order to test the performance of our method in a low overlap ratio, we placed the far point for the source point cloud and target point cloud independently. The other pre-processing steps are the same as those in the third experiment involving Gaussian noise. Table~\ref{table4} displays the results, indicating that our method continues to outperform other methods in terms of performance. Additionally, a qualitative comparison of the registration results is presented in Fig.~\ref{figure6}(d).

\subsection{Evaluation on Stanford 3D Scan}
To assess the generalizability, we perform experiments using the Stanford 3D Scan dataset. As this dataset contains only 10 real scans, we utilized the ModelNet40 trained model without fine-tuning. Some examples are shown in Fig.~\ref{figure7}.

\subsection{Evaluation on 7Scenes}
\begin{figure*}
	\centering
	\includegraphics[width=0.95\textwidth]{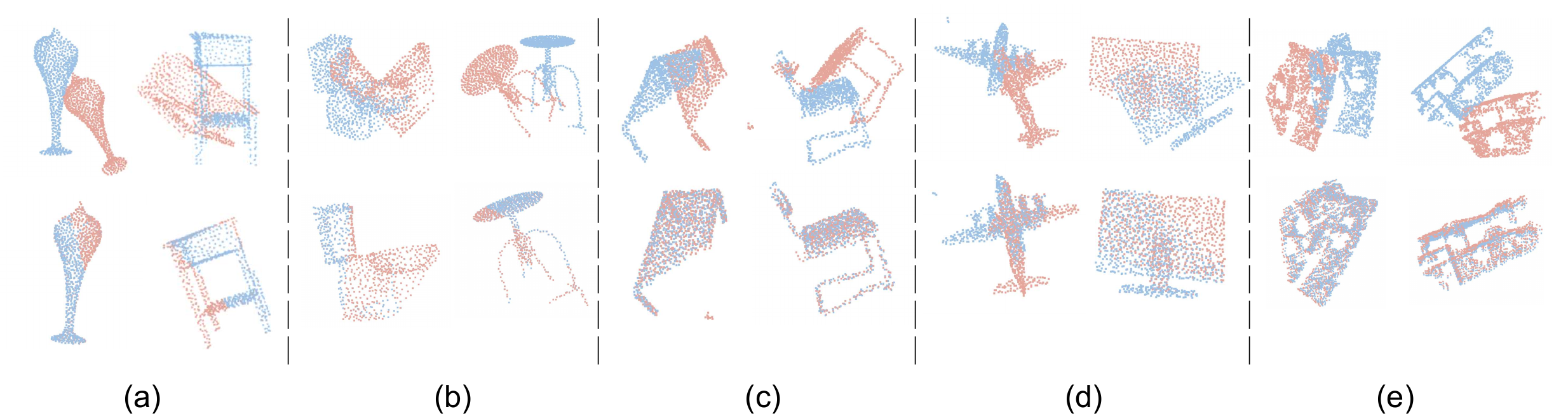}  
	\caption{Qualitative results. (a) Unseen shapes on ModelNet40. (b) Unseen categories on ModelNet40. (c) Gaussian noise on ModelNet40. (d) Gaussian noise with lower overlap on ModelNet40. (e) 7Scenes. (top: initial positions, bottom: registration results)}
	\label{figure6}
	\vspace{-.1cm}
\end{figure*}

\begin{figure}[t]
	\centering
	\includegraphics[width=0.48\textwidth]{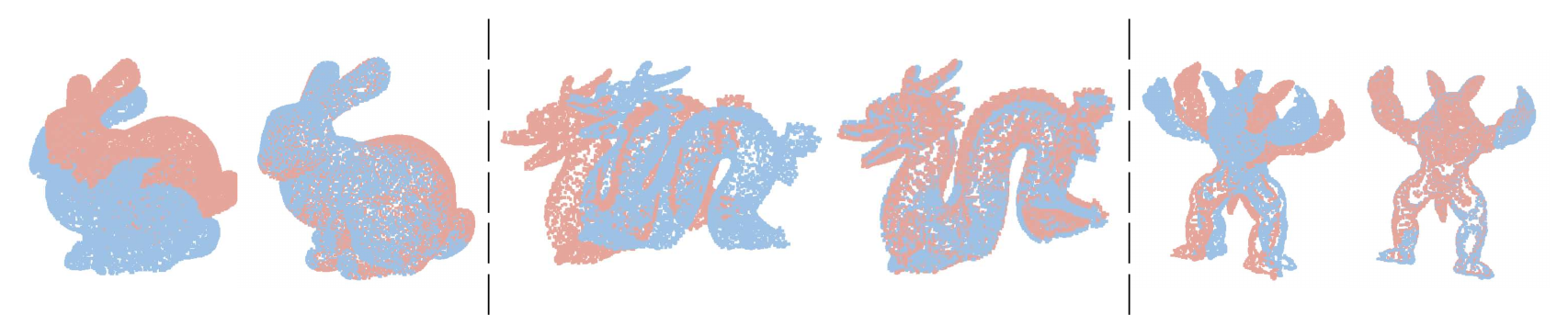} 
	\vspace{-.3cm}
	\caption{Qualitative results on Stanford 3D Scan.
	}
	\label{figure7}
	\vspace{-.3cm}
\end{figure}

We conduct a comparative evaluation on the real-world dataset 7Scenes. Our model is trained on 6 categories (\emph{Chess}, \emph{Fires}, \emph{Heads}, \emph{Pumpkin}, \emph{Stairs} and \emph{Redkitchen}) and tested on the remaining category (\emph{Office}). We resample the source point clouds to 2,048 points and apply rigid transformation to generate the target point clouds, we then downsample the point clouds to 1,536 points to generate the partial data. From Table~\ref{table5}, one can see that our method achieves outstanding performance on real-world scenes. Fig.~\ref{figure6}(e) depicts some examples of 7Scenes.
\begin{table}[]
    \small
    \centering
    \setlength\tabcolsep{8pt} 
    \caption{The registration result on Gaussian noise in ModelNet40.}
    \resizebox{\linewidth}{!}{%
        \begin{tabular}{lcccccc}
        
            \toprule
            Method     &RMSE(\textbf{R})       &MAE(\textbf{R})  &RMSE(\textbf{t})  &MAE(\textbf{t})        
            \\ 
            \midrule
            ICP~\cite{besl1992method}        & 35.077&25.562&0.2925&0.2491 \\
            PointNetLK~\cite{aoki2019pointnetlk} &18.926&8.944&0.0647&0.0423  \\
            DCP~\cite{wang2019deep} &6.925&4.487&0.0242&0.0187 \\
            PRNet~\cite{Wang_Solomon_2019} &4.323&2.196&0.0195&0.0140 \\
            IDAM~\cite{li2020iterative}&3.721&1.855&0.0232& 0.0118  \\
            OMNet~\cite{xu2021omnet} & 2.373&0.948&0.0168&0.0086 \\
            FINet~\cite{xu2022finet} &\underline{1.706}&\underline{0.937}&0.0124 &0.0084 \\ 
            VRNet~\cite{zhang2022vrnet} &3.615&1.637& \underline{0.0101}&\underline{0.0063} \\
            \midrule
            Ours & \textbf{1.472}&\textbf{0.632}& \textbf{0.0058}& \textbf{0.0036}
            \\ 
            \bottomrule
        \end{tabular}%
    }
    \vspace{-0.3cm}
    \label{table3}
\end{table}

\begin{table}[]
    \small
    \centering
    \setlength\tabcolsep{8pt} 
    \caption{The registration result on Gaussian noise with lower overlap in ModelNet40. Besides, VRNet does not provide the results on Gaussian noise with lower overlap.}
    \resizebox{\linewidth}{!}{%
        \begin{tabular}{lcccccc}
        
            \toprule
            Method     &RMSE(\textbf{R})       &MAE(\textbf{R})  &RMSE(\textbf{t})  &MAE(\textbf{t})        
            \\ 
            \midrule
            ICP~\cite{besl1992method}        &64.412&46.943&0.9422&0.8571 \\
            PointNetLK~\cite{aoki2019pointnetlk} &38.991&21.262&0.2451&0.1429  \\
            DCP~\cite{wang2019deep} &9.932&6.821&0.0969&0.0724 \\
            PRNet~\cite{Wang_Solomon_2019} &8.116&4.942&0.0871&0.0498 \\
            IDAM~\cite{li2020iterative}&9.603&5.296&0.1006& 0.0544  \\
            OMNet~\cite{xu2021omnet} &\underline{4.972}&3.567&0.0524&0.0381 \\
            FINet~\cite{xu2022finet} & 5.059&\underline{2.894}&\underline{0.0360}&\underline{0.0273} \\ 
            \midrule
            Ours & \textbf{4.341}&\textbf{2.279}& \textbf{0.0214}& \textbf{0.0104}
            \\ 
            \bottomrule
        \end{tabular}%
    }
    \vspace{-0.3cm}
    \label{table4}
\end{table}

\begin{figure}[t]
	\centering
	\includegraphics[width=0.45\textwidth]{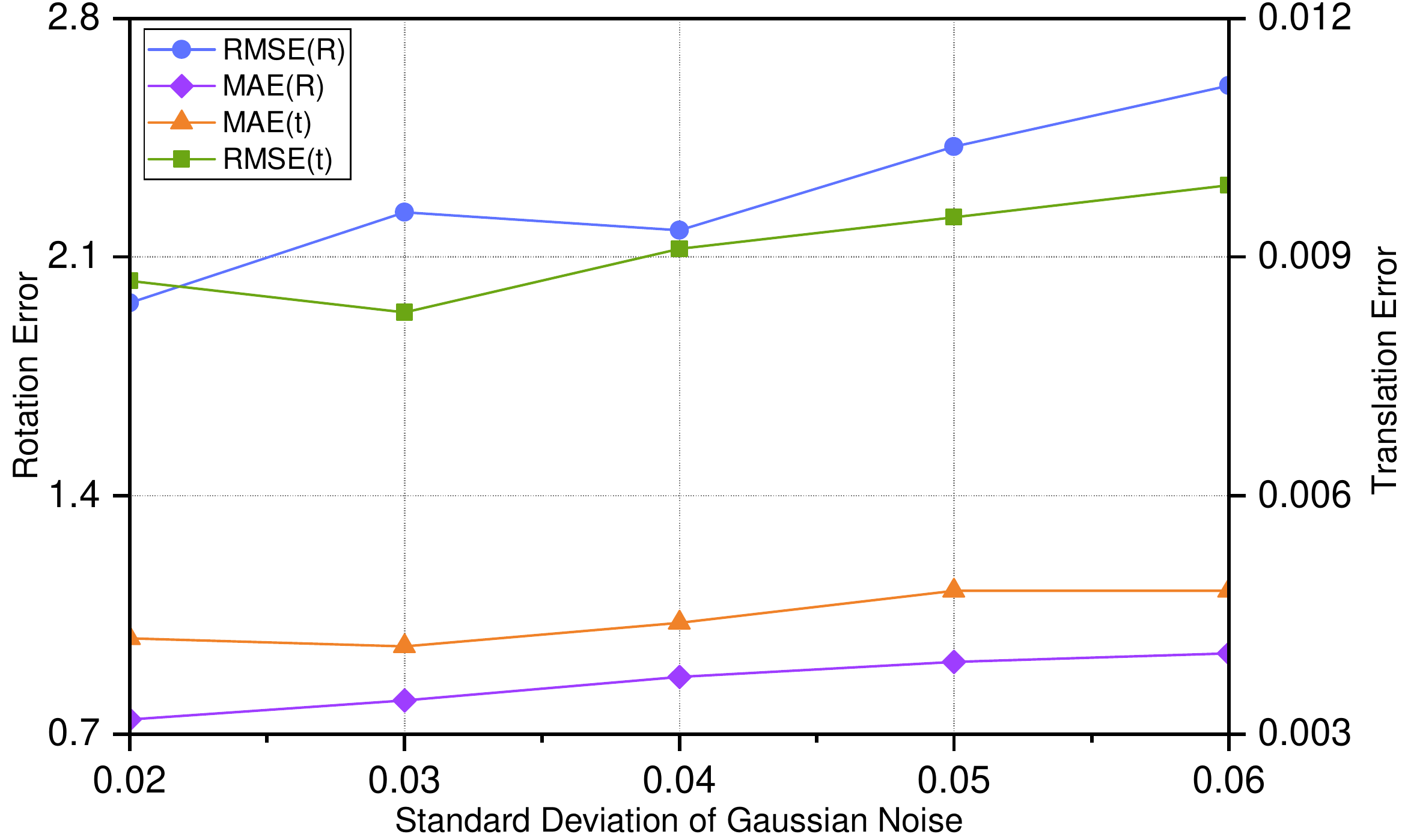} 
	\vspace{-.3cm}
	\caption{Errors of our method under different noise levels.
	}
	\label{figure8}
	\vspace{-.3cm}
\end{figure}

\begin{figure}[t]
	\centering
	\includegraphics[width=0.45\textwidth]{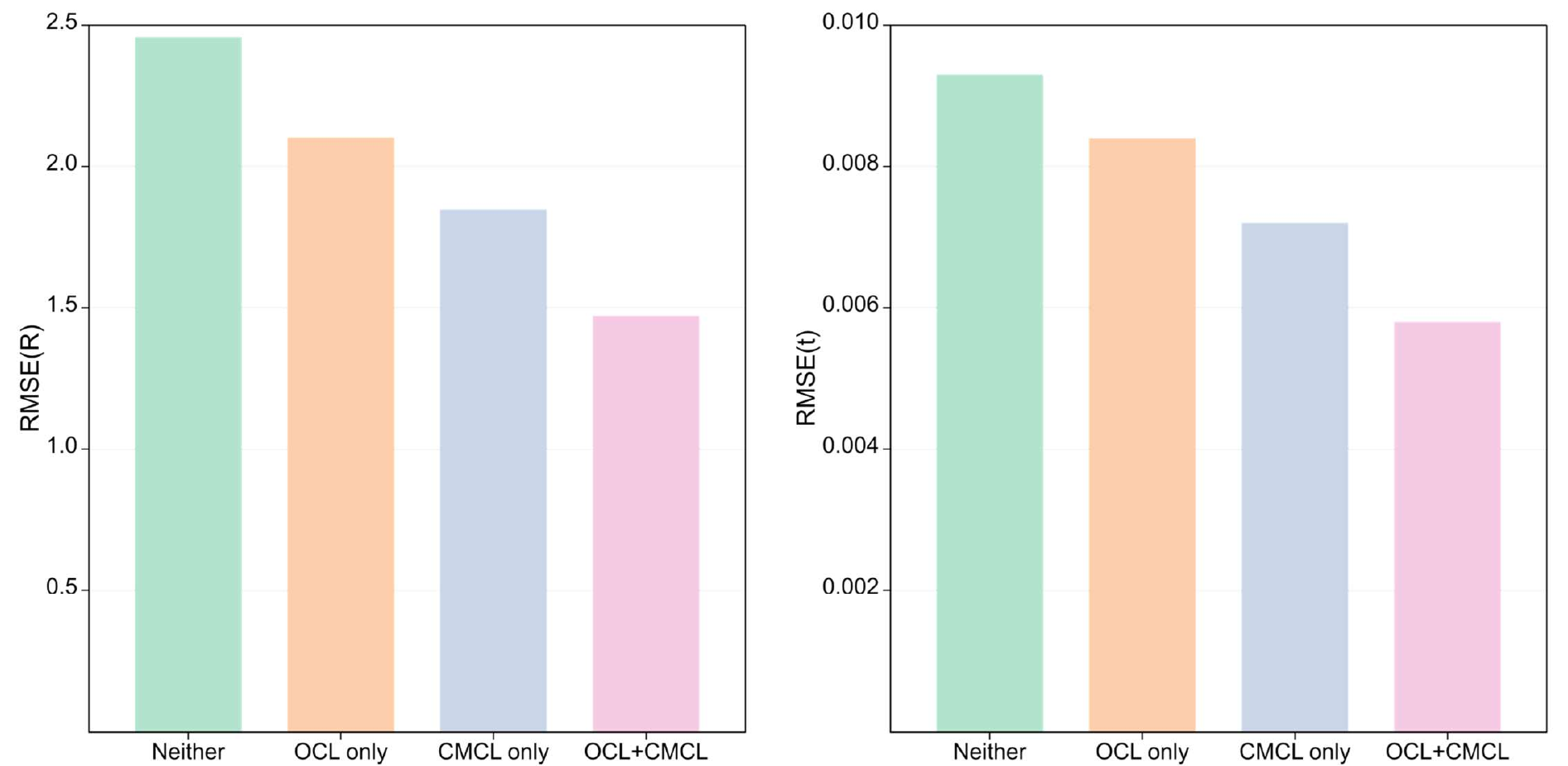} 
	\vspace{-.3cm}
	\caption{Impact of the joint contrastive learning(OCL+CMCL) when compared to individual overlapping contrastive learning(OCL only), individual cross-modal contrastive learning(CMCL only) and neither is used(Neither).  }
	
	\label{figure9}
	\vspace{-.3cm}
\end{figure}

\begin{table}[]
    \small
    \centering
    \setlength\tabcolsep{8pt} 
    \caption{The registration result on 7Scenes. Besides, VRNet does not provide the results on 7Scenes.}
    \resizebox{\linewidth}{!}{%
        \begin{tabular}{lcccccc}
        
            \toprule
            Method     &RMSE(\textbf{R})       &MAE(\textbf{R})  &RMSE(\textbf{t})  &MAE(\textbf{t})        
            \\ 
            \midrule
            ICP~\cite{besl1992method}       & 10.416 & 6.194 & 0.1979 & 0.0173 \\
            PointNetLK~\cite{aoki2019pointnetlk} & 4.055 & 2.908 & 0.0325 & 0.0092 \\
            DCP~\cite{wang2019deep} & 6.742 & 4.195 & 0.0376 & 0.0213 \\
            PRNet~\cite{Wang_Solomon_2019} & 2.915 & 1.143 & 0.0142 & 0.0097 \\
            IDAM~\cite{li2020iterative} & 8.594 & 5.761 & 0.0329 & 0.0231 \\
            OMNet~\cite{xu2021omnet} & \underline{1.449} & \underline{0.836} & \underline{0.0071} &\underline{0.0047} \\
            FINet~\cite{xu2022finet} & 1.782 & 0.903 & 0.0094 & 0.0051 \\ 
            \midrule
            Ours & \textbf{0.804} & \textbf{0.488} & \textbf{0.0032} & \textbf{0.0018} 
            \\ 
            \bottomrule
        \end{tabular}%
    }
    \vspace{-0.3cm}
    \label{table5}
\end{table}

\begin{table}[t]
    \caption{Ablation studies of each component.}
    \resizebox{\linewidth}{!}{%
    \begin{tabular}{l|cccc|cccc}
    \toprule
    \# & TF & CMD & MCL & MP 
    &RMSE(\textbf{R}) & MAE(\textbf{R}) & RMSE(\textbf{t})  &MAE(\textbf{t}) \\
    \midrule
    
    1  &-  &- &- &-
     & 3.565 & 1.529 & 0.0194 & 0.0081\\ 
    
    2  &\checkmark &- &- &-
    &3.106 & 1.299 &0.0137 &0.0064\\ 

    3  &\checkmark &\checkmark &- &-
    &2.537 & 1.026 &0.0106 &0.0069\\ 

    4  &\checkmark &\checkmark &\checkmark &-
    &1.708 & 0.713 &0.0066 &0.0039\\
    
    \midrule
    5  & \checkmark & \checkmark &\checkmark &\checkmark
     & \textbf{1.472} & \textbf{0.632} & \textbf{0.0058} & \textbf{0.0036}\\
    \bottomrule
    \end{tabular}
    }
    \label{table6}
\end{table}

\begin{table}[t]
    \caption{Ablation studies of different iteration times.}
    \resizebox{\linewidth}{!}{%
    \begin{tabular}{lcccccccccccc}
        \toprule
         n      & RMSE(\textbf{R}) & MAE(\textbf{R}) & RMSE(\textbf{t}) & MAE(\textbf{t}) & Inference time \\
        \midrule
        2        & 1.972          & 0.748         & 0.0084          & 0.0041      & \textbf{133.0}        \\
        3           & \textbf{1.472}       & \textbf{0.632}   & \textbf{0.0058}  & \textbf{0.0036}   & 151.1         \\
        4      & 1.564         & 0.664          & 0.0063       & 0.0037  & 165.9\\
        5         & 1.556          & 0.668         & 0.0063       & 0.0038      & 179.9  \\
        
         \bottomrule
    \end{tabular}%
    }
    
    \vspace{-0.37cm}
	\label{table7}
\end{table}

\begin{table}[t]
    \caption{Ablation studies of the Correspondences Search module.}
    \resizebox{\linewidth}{!}{%
    \begin{tabular}{lcccccccccccc}
        \toprule
         Method      & RMSE(\textbf{R}) & MAE(\textbf{R}) & RMSE(\textbf{t}) & MAE(\textbf{t}) \\
        \midrule
        PCRNet~\cite{sarode2019pcrnet}        & 14.811          & 5.720         & 0.0525          & 0.0274         \\
        IDAM~\cite{li2020iterative}          & 2.790       & 1.079   & 0.0104  & 0.0051      \\
        \midrule       
        Ours       & \textbf{1.472}       & \textbf{0.632}   & \textbf{0.0058}  & \textbf{0.0036} \\

         \bottomrule
    \end{tabular}%
    }
    
    \vspace{-0.37cm}
	\label{table8}
\end{table}

\begin{figure}[t]
	\centering
	\includegraphics[width=0.42\textwidth]{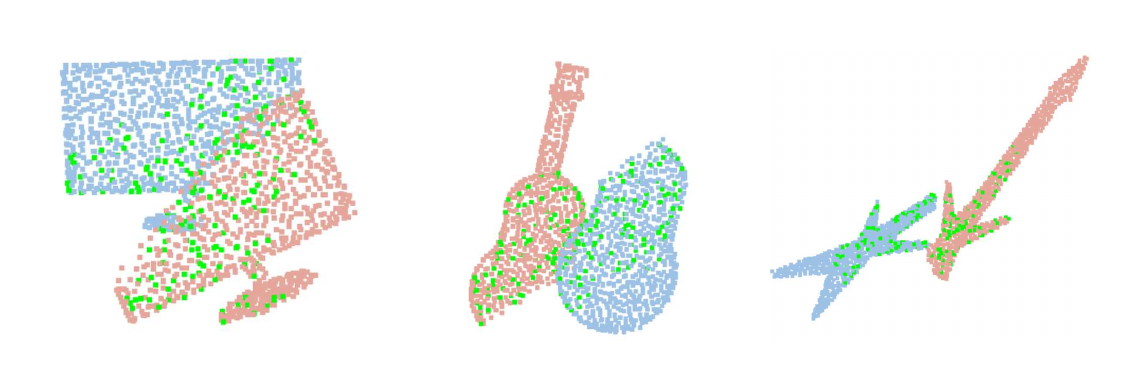} 
	\vspace{-.3cm}
	\caption{Visualization of the keypoints.(blue: source point cloud, red: target point cloud, green: keypoints.)
	}
	\label{figure10}
	\vspace{-.3cm}
\end{figure}

\subsection{Robustness Analysis}
To demonstrate the robustness of our method, we train and test our models using varying degrees of noise, as illustrated in Fig.~\ref{figure8}. We accomplish this by introducing noise that is sampled from $N(0, \sigma^{2})$ and then clipped within the range of $[-0.05, 0.05]$, where the deviation $\sigma \in [0.02, 0.06]$. Notably, our method consistently achieved comparable performance across varying noise levels.

\subsection{Ablation Studies}
In this section, we present the results of several ablation experiments on the Gaussian noise to demonstrate the effectiveness of our components and settings.
In our baseline setting, we do not use any cross-modal data or mask prediction.
As shown in Table~\ref{table6}, we can find that all the components improve the performance.

\emph{Transformer Fusion (TF) and Cross-Modal Data (CMD).}
The Transformer Fusion module is responsible for combining 3D point cloud features with 2D image features in a logical manner, resulting in the creation of multimodal hybrid features. Comparing Row 1 and Row 3 in Table~\ref{table6}, it becomes apparent that the inclusion of cross-modal image information can significantly improve the accuracy of registration. 
To demonstrate the importance of cross-modal image information, we remove cross-modal information (CMD) while retaining Transformer Fusion (TF), as shown in Row 2 of Table~\ref{table6}, where the results drop dramatically after removing cross-modal information.

\emph{Multiple Contrastive Learning (MCL).}
Comparing Row 3 with Row 4 in Table~\ref{table6}, we can observe that applying contrastive learning loss can lead to significant improvements. This is because overlapping contrastive learning highlights the features of overlapping points, and cross-modal contrastive learning enables 3D point cloud features to correspond with 2D image features, facilitating feature fusion. 
Fig.~\ref{figure9} graphically illustrates that each contrastive learning strategy has a positive effect on the registration results.

\emph{Mask Prediction (MP).}
Comparing Row 4 with Row 5 in Table~\ref{table6}, it becomes evident that the Mask Prediction module has a beneficial impact. This module is employed to mitigate the undesirable effects on the registration task that are akin to non-overlapping points. As shown in Fig.~\ref{figure10}, the green points represent the keypoints extracted by mask prediction (MP), these points are more inclined to edge points and overlapping points, which are more helpful for registration tasks.

\emph{Correspondences Search and Iteration times.}
We compare the performance of our method for different iteration times. Specifically, we set $\mathrm{n}$ to 2, 3, 4, and 5. The ablation studies of different iteration times are presented in Table~\ref{table7}. To achieve a balance between efficiency and performance, we set $\mathrm{n}$ to 3 for all experiments conducted. 
Additionally, in order to verify the effectiveness of the Correspondences Search module, we replace it with other similar structures, and the experimental results are shown in Table~\ref{table8}, in which PCRNet~\cite{sarode2019pcrnet} uses direct regression to obtain the rigid transformation, and IDAM~\cite{li2020iterative} is similar to ours but does not compute the geometric information separately from the higher-level features. As can be seen from the table, the Correspondences Search module designed by us can get the optimal performance.

\section{CONCLUSIONS}

We present the CMIGNet, a novel method that utilizes cross-modal information for point cloud registration. Previous methods based on global features are prone to incorrectly treating outlier correspondences with similar local structures as inlier correspondences. However, our method perceives the global shape by learning cross-modal information to achieve more accurate registration.
Specifically, we propose two contrastive learning strategies: overlapping contrastive learning to highlight overlapping point features and cross-modal contrastive learning to achieve 2D-3D correspondences. We then use an attention mechanism to achieve information interaction and feature fusion. We also develop a new mask prediction method to select keypoints in the point cloud.
Extensive experiments on the ModelNet40, Stanford 3D Scan, and 7Scenes benchmarks demonstrate that our method can achieve outstanding performance.

\addtolength{\textheight}{-3cm}   

\bibliographystyle{IEEEtran}
\balance

\end{document}